\documentclass[
]{ceurart}

\usepackage{listings}
\usepackage{subcaption} 
\usepackage{enumitem}

\lstdefinelanguage{json}{
    morestring=[b]",
    morestring=[d]',
    morecomment=[l]{//},
    morekeywords={true,false,null},
    sensitive=false,
}

\lstdefinestyle{compactjson}{
    language=json,
    basicstyle=\ttfamily\footnotesize,
    frame=single,
    backgroundcolor=\color{gray!5},
    breaklines=true,
    showstringspaces=false,
    aboveskip=0pt,
    belowskip=0pt,
    framexleftmargin=4pt,
    framextopmargin=1pt,
    framexbottommargin=1pt,
    xleftmargin=0pt,
    xrightmargin=0pt,
    tabsize=2
}

\begin{document}

\copyrightyear{2025}
\copyrightclause{Copyright for this paper by its authors.
  Use permitted under Creative Commons License Attribution 4.0
  International (CC BY 4.0).}

% \conference{Joint proceedings of KBC-LM and LM-KBC @ ISWC 2025}

\title{Towards Temporal Knowledge-Base Creation for Fine-Grained Opinion Analysis with Language Models}

% \tnotemark[1]
% \tnotetext[1]{You can use this document as the template for preparing your
%   publication.}

\author[1]{Gaurav Negi}[%
orcid=0000-0001-9846-6324,
email=gaurav.negi@insight-centre.org 
% url=https://github.com/gauneg,
]
% \cormark[1]
% \fnmark[1]
% \address[2]{Joint Institute for Nuclear Research}

\author[1]{Atul Kr. Ojha}[%
orcid=0000-0002-9800-9833,
email=atulkumar.ojha@insight-centre.org,
url=https://github.com/shashwatup9k,
]
% \fnmark[1]
% \address[3]{Vrije Universiteit Amsterdam, De Boelelaan 1105, 1081 HV Amsterdam, The Netherlands}

\author[1]{Omnia Zayed}[%
orcid=0000-0002-8357-8734,
email=omnia.zayed@insight-centre.org,
url=https://www.insight-centre.org/our-team/omnia-zayed/,
]

\author[1]{Paul Buitelaar}[%
orcid=0000-0001-7238-9842,
email=paul.buitelaar@insight-centre.org,
url=https://research.universityofgalway.ie/en/persons/peter-paul-buitelaar,
]

\address[1]{Insight SFI Research Ireland Centre for Data Analytics, University of Galway}

% \fnmark[1]
% \address{University of Galway}

%% Footnotes
% \cortext[1]{Corresponding author.}
% \fntext[1]{These authors contributed equally.}
\conference{Joint proceedings of KBC-LM and LM-KBC @ ISWC 2025}
\maketitle
\begin{abstract}
We propose a scalable method for constructing a temporal opinion knowledge base with large language models (LLMs) as automated annotators. Despite the demonstrated utility of time-series opinion analysis of text for downstream applications such as forecasting and trend analysis, existing methodologies underexploit this potential due to the absence of temporally grounded fine-grained annotations. Our approach addresses this gap by integrating well-established opinion mining formulations into a declarative LLM annotation pipeline, enabling structured opinion extraction without manual prompt engineering. We define three data models grounded in sentiment and opinion mining literature, serving as schemas for structured representation. We perform rigorous quantitative evaluation of our pipeline using human-annotated test samples. We carry out the final annotations using two separate LLMs, and inter-annotator agreement is computed label-wise across the fine-grained opinion dimensions, analogous to human annotation protocols. The resulting knowledge base encapsulates time-aligned, structured opinions and is compatible with applications in Retrieval-Augmented Generation (RAG), temporal question answering, and timeline summarisation.

\end{abstract}

\section{Introduction}
The analysis of public opinions on the internet has been an indispensable field of research for real-time impactful applications. With the advent and the success of multiple platforms, the internet has become a reliable source of low-latency public opinion at a significant scale. This continuous flux of opinionated text data has proven its usability not just in analytics but also in prediction and forecasting research. The value of information provided by analysis of timed social media data is well-founded; however, the utilisation of temporal data collected for these studies remains underutilised due to the lack of annotation following the existing fine-grained opinion formulations. 

There has been a considerable evolution in the field of fine-grained opinion analysis, which has led to many theoretical and practical formulations of opinions and sentiment \cite{liu2017many, absa_survey_2022, DBLP:conf/semeval/PontikiGPPAM14}. These fine-grained methods help get a clearer picture of not just the sentiment polarity (positive, negative, neutral) but also the contextual elements of the aforementioned sentiments. The use of these opinion formalisms has not been explored in the temporal analysis of opinions over social media posts. In our methodologies, we utilise these well-researched formulations to describe the schema for describing opinions in the knowledge base.

In the era of large language models (LLMs), Retrieval-Augmented Generation (RAG) has demonstrated significant success specifically for fact-driven, knowledge dependant tasks \cite{rag_2020}. Furthermore, the consideration of temporal aspect while retrial \cite{ts_rag_2025} have shown to benefit LLM based forecasting systems. There is also a strong argument for having a well-structured knowledge base to improve the performance of RAG systems \cite{Zhu_Guo_Cao_Li_Gong_2024, graph_rag_survey_2025}. This serves as our motivation for creating a large-scale, well-structured knowledge base for temporal public opinions. 

The manual and periodic annotation of social media data to accommodate unforeseeable events, including outbreaks, financial events, and political events, is highly impractical and expensive. To overcome this challenge, we introduce a framework for automatic opinion annotation of the time-stamped data collected from social media. Thus, resulting in the creation of a temporal, structured knowledge base for opinions. LLMs serve as the backbone of our annotation pipeline as they are a competent few-shot inferer with a task-agnostic architecture \cite{brown_et_al_2020}. 
% Their use in reasoning and judgments \cite{llm_zeroshot_reasoner_2022, llm_judge_23} is well documented.

The key contributions of this work are as follows:
\begin{itemize}
    \item \textbf{Data Model Definition} We describe three data models based on a well-researched area of opinion mining and sentiment analysis in text. These serve as the schema for describing opinions in a knowledge base.
    
    \item \textbf{Declarative Structure-Conforming LLM Based Annotation}: We apply declarative methods for reliably annotating datasets with LLMs, which aims to reduce the heavy reliance on manual prompt design and engineering. This method removes the manual prompt construction and post-processing of LLM results required to extract results adhering to the desired structures specified by data models. 

    \item \textbf{Multi-Schema Opinion Knowledge-Base}: A knowledge base creation that has structured opinions for the data collected from social media platforms. This knowledge base has high potential for practical applications in time-series analysis, RAG Question Answering, and Abstractive and Extractive Timeline Summarisation. The annotated dataset is available publicly at \url{https://github.com/ANON-1221/KBC-LM-Temporal-Opinions}.
\end{itemize}

\section{Related Work}

\subsection{Time Series Analysis of Social Media Opinion}
In the early days of social media advent, \citet{connor_2010} showed the potential of social media sentiment in replacing telephone-based polling. There have been numerous well-received attempts at modelling social media opinions over time \cite{topic_sentiment_evolve_2014, tracking_setiment_time_series_2016, sentiment_bursts_social_2018, topic_senti_evol_2020}. The temporal assessment of opinionated social media text has also proven valuable in monitoring various public health dimensions. During the Covid-19 pandemic, both in the short-term \cite{sentiment_dynamics_covid_2020, covid_topics_2020} and its long-term effects \cite{sentiment_alteration_global_2022} were inferred with social media text analysis. There has been research in utilising the social media data for public mental health monitoring \cite{low2020natural} and stock market forecast \cite{lee2023stockemotions}.

\subsection{Subjective Knowledge Bases}
Subjective knowledge bases are not a new concept. In neuro-symbolic learning, they continue to play a role by bridging the parametric learning of LLMs with symbolic representations. SenticNet \cite{senticnet_2015} is a knowledge base that offers phrase level affective information to aid in subjective tasks. Similarly, OpineDB \cite{subdb_2019} has opinionated reviews for hotels and restaurants for aiding opinion mining systems. These existing datasets exhibit an exemplary efficacy in establishing a rigorous knowledge base for capturing the subjectivity, specifically sentiment. However, they do not provide a temporal axis in the knowledge base which plays a key role in analysis of dynamics of sentiment/opinions.
\subsection{Opinion  Formulations.}
Opinion\footnote{Unless stated otherwise, we use the term opinion as a broad concept that covers sentiment and its associated information, such as opinion target and the person who holds the opinion, and use the term sentiment to mean only the underlying positive, negative or neutral polarity implied by opinion.} mining and sentiment analysis has been well explored in natural language processing. Various fine-grained formulations of opinion mining that fall under the subdomain of Aspect-Based Sentiment Analysis (ABSA) offer multiple alternatives for data models for a temporal knowledge base.   ABSA has evolved from feature-based summarisation \cite{hu_mining_2004,zhuang_movie_2006,ding2008liu} and the foundational work on opinion mining by \citet{DBLP:books/sp/mining2012/LiuZ12}, which involves extracting and summarising opinions on features (attributes/keywords). The downstream tasks that spun out of the ABSA research space can be specified into the following categories based on the opinion facets they address: \textit{Opinion Aspect Co-extraction} \cite{qiu_opinion_2011,liu_opinion_coextratcion_2013,liaspect,wang_coupled_2017}, \textit{Aspect Sentiment Triple Extraction (ASTE)} \cite{aste_ote_mtl,aste_jet,grid_at_op_sentiment_2020}, \textit{Aspect-Category-Opinion-Sentiment Quadruple (ACOS/ASQP) extraction} \cite{acos_extract_classify, mvp_2023,acos_gen_nat_scl_bart}. 

Other than ABSA, formulations of fine-grained opinion include \citet{barnes_structured_2021, barnes_flaunt_senti_2021}. They perform \textit{Structured Sentiment Analysis} by extracting sentiment tuples as dependency graph parsing, where the nodes are spans of sentiment holders, targets, and expressions, and the arcs are the relations between them. \citet{negi_uoc_2025} introduce Unified Opinion Concepts (UOC) ontology to integrate opinions within their semantic context, expanding on the expressiveness by conceptualising multi-faceted \cite{liu2017many} opinion into an ontological form. 

We use these formulations to create data models for the opinion schema of the annotated knowledge base we created.

\paragraph{Data Annotation Using Large Language Models} 
Data Annotation for training computational models has always been a human-driven field. However, due to the success of LLMs in generalised use, there have been investigations in their use for data annotation \cite{llm_annotation_2024}. Their use as argumentation \cite{llm_argue_annot_2024} and span annotation\cite{llm_span_annot_2025} has been found to have promising results. 

A substantial part of our work also deals with annotation using LLMs, but unlike existing work, we focus specifically on different opinion annotation for creating a time-aware knowledge base. 

\section{Data Models and Resources}
\label{sec:data_models}

\subsection{Preliminaries}
We use the facets of opinions and sentiment to describe the classes of the data models that we formalise based on the existing literature. In order to fully understand the date models, we briefly describe the concepts widely used in fine-grained opinion analysis:
\begin{itemize}
    
    \item \textbf{Sentiment}: This concept encapsulates the underlying
feelings expressed in an opinion. It is a composite concept that often encapsulates descriptive properties, namely sentiment polarity (positive, negative, or neutral), sentiment intensity (strong, average, or weak), and sentiment expression in the text. However, in the existing literature, sentiment has often been used synonymously with sentiment polarity. Consider the following Example:
    \vspace{-10pt}
    \paragraph{Example} \textit{I had hoped for better battery life, as it had only about 2-1/2 hours doing heavy computations (8 threads using 100\% of the CPU)}

    Here, the sentiment is specified by (i) Sentiment Expression: \textit{``hoped for better''}, (ii) Sentiment Polarity: \textit{negative} and (iii) Sentiment Intensity: \textit{average}
    
    \item \textbf{Target}: The subjective information on which an opinion is expressed. In some formulations, it is expressed as a span of text; in others, it may be an abstract concept formed by combining aspect elements and a coarse-grained Target Entity. 

    In the Example, the opinion is expressed regarding the Target entity: \textit{Battery}.
    
    \item \textbf{Aspect}: Aspect describes the part and attribute of the Target Entity on which the sentiment is expressed. It is made up of an instantiation explicitly in the text called an aspect term and a more coarse-grained property called a category, which expresses the attribute of the entity towards which the opinion is specifically directed. 

    The Aspect of opinion in the Example can be expressed into the Aspect Term: \textit{``battery life''} and Aspect Category: \textit{Operation\_Performance}

    \item \textbf{Holder}: A holder is the explicit individual or organisation that is expressing the opinion. An opinion is expressed in the first person in the Example, making the span \textit{``I''} the holder.
    
    \item \textbf{Qualifier}: A Qualifier refines the scope or applicability of an opinion, delineating the group or subgroup to which the opinion pertains. For instance, in the above Example, the only subgroup of people affected is those performing heavy computations. Therefore \textit{``doing heavy computations''} is the Qualifier.

    \item \textbf{Reason}: It represents an opinion's justification or the underlying cause. The Reason for the opinion in the Example specifically addresses the battery issues, i.e. \textit{``it had only about 2-1/2 hours''}

\end{itemize}

% \begin{figure}[htbp]
%   \centering
%   \begin{subfigure}{0.48\linewidth}
%     \centering
%     \includegraphics[width=\linewidth]{images/acos.pdf}
%     \caption{Structured Sentiment Analysis (SSA) \cite{structured_sentiment_barns_2022}}
%   \end{subfigure}
%   % \hfill
%   \hspace{0.02\linewidth}
%   \begin{subfigure}{0.48\linewidth}
%     \centering
%     \includegraphics[width=\linewidth]{images/ssa.pdf}
%     \caption{Aspect-Category-Opinion-Sentiment (ACOS) \cite{acos_extract_classify}}
%   \end{subfigure}\par\medskip
%   \begin{subfigure}{\linewidth}
%     \centering
%     \includegraphics[width=.98\linewidth]{images/uoce.pdf}
%     \caption{Unified Opinion Concepts (UOC) \cite{negi_uoc_2025}}
%   \end{subfigure}
%   \caption{Opinion data models for the annotation of the Temporal Knowledge Base}
%   \label{fig:variants}
% \end{figure}

\subsection{Data Models}
The concepts described in preliminaries have been composed and used for fine-grained opinion formulations. We use three accepted and fine-grained formulations to describe data models for opinion representation, which are described subsequently. The availability of human-annotated datasets for these formulations played a decisive role in our data modelling process. Even though our annotation pipeline does not fine-tune LLMs, it still relies on human annotations for configuring declarative LLM generative methods and their evaluations.
% \begin{itemize}
    
    \paragraph{Aspect-Category-Opinion-Sentiment}: This data model is based on named Aspect-Category-Opinion-Sentiment (ACOS) Quadruple Extraction, to extract aspect, aspect category, opinion and sentiment as quadruples in text and provide full support for aspect-based sentiment analysis with implicit aspects and opinions \footnote{Opinion in ACOS task refers to the span that expresses sentiment.}. The primary focus of this formulation is the fine-grained analysis of the opinion target.

    \begin{figure}[htbp]
    \centering
    \includegraphics[width=0.7\linewidth]{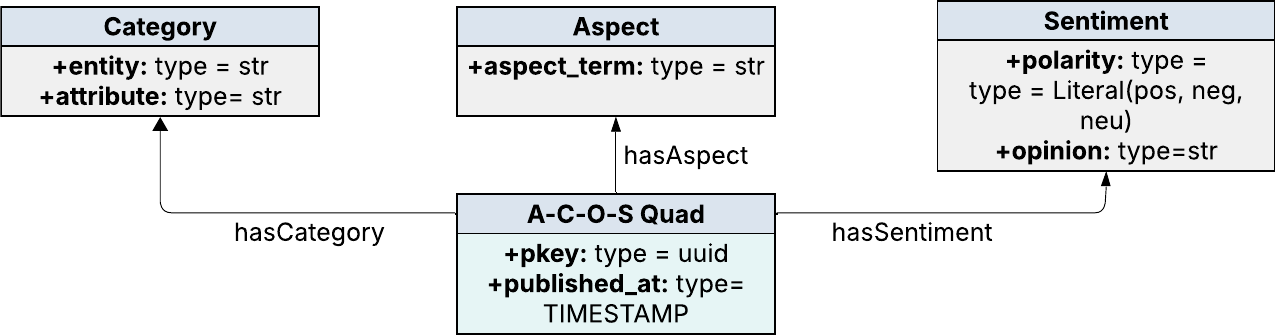}
    \caption{Aspect-Category-Opinion-Sentiment (ACOS) \cite{acos_extract_classify}}
    \label{fig:acos}
    \end{figure}
    
    \paragraph{Structured Sentiment Analysis}: The data model is built upon the structured sentiment formulation aimed at performing opinion tuple extraction as dependency graph parsing, where the nodes are spans of sentiment holders, targets and expressions. The interconnection between them shows the existence of relationships among these components. The primary focus of this formulation is the fine-grained nature of expressed sentiment.

    \begin{figure}[htbp]
    \centering
    \includegraphics[width=0.7\linewidth]{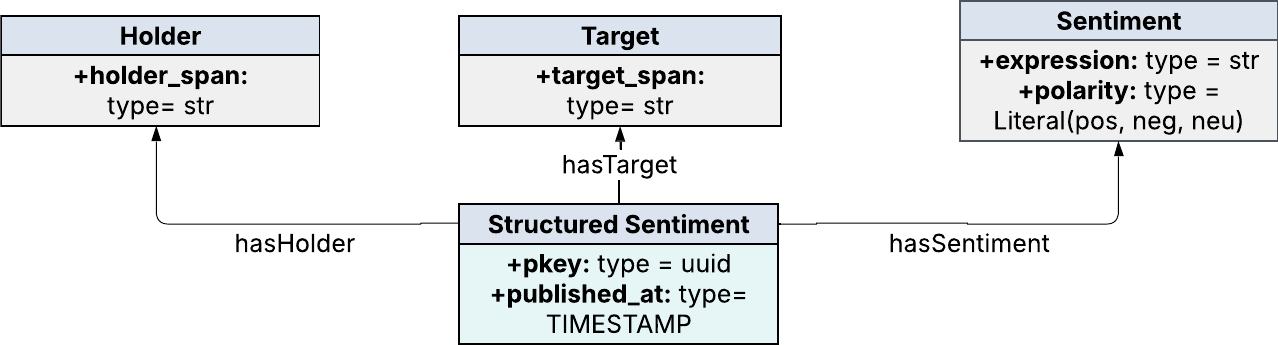}
    \caption{Structured Sentiment Analysis (SSA) \cite{structured_sentiment_barns_2022}}
    \label{fig:ssa}
    \end{figure}
    
\begin{figure}[htbp]
    \centering
    \includegraphics[width=0.8\linewidth]{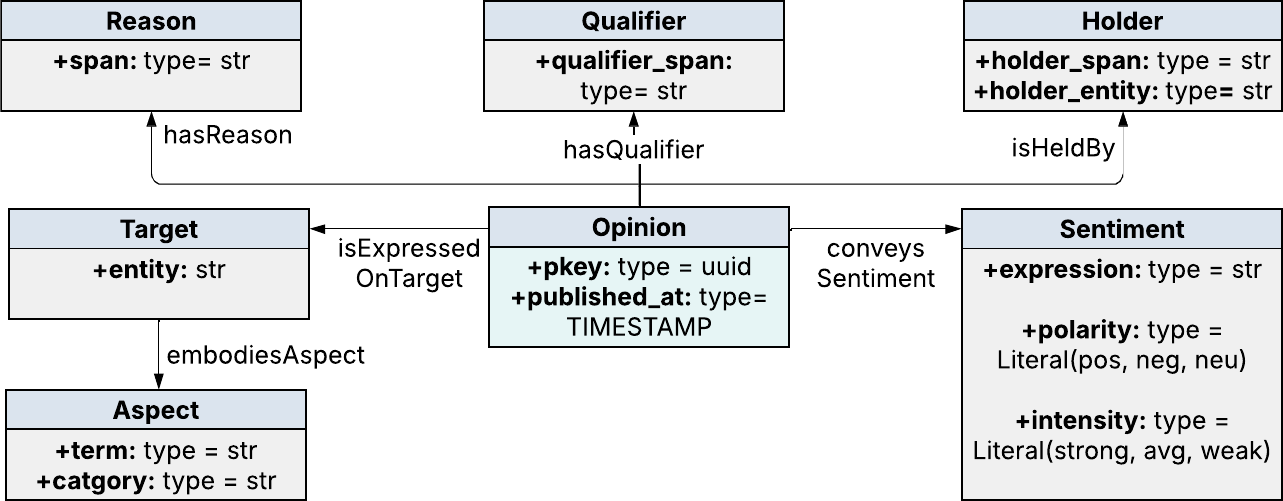}
    \caption{Unified Opinion Concepts \cite{negi_uoc_2025}}
    \label{fig:uoc}
    \end{figure}

    \paragraph{Unified Opinion Concepts}: This data model is based on the UOC ontology that bridges the gap between the semantic representation of opinion across different formulations. It is a unified conceptualisation based on the facets of opinions studied extensively in NLP and semantic structures described through symbolic descriptions. As seen in Figure \ref{fig:uoc}, it brings together the elements of SSA and ACOS formulation.

\vspace{-20pt}
\subsection{Datasets}
We use two types of datasets in our work: (i) Data Model Datasets, (ii) Temporal Knowledge-Base Dataset. We use the former one as source datasets to train and evaluate our pipeline for the extraction of opinions adhering to the specified data models (see Table \ref{tab:dmodel_dsets}). The latter one is used for creating the temporal knowledge bases for opinions are described below along with quantitative details in Table \ref{tab:tdatasets}:

\begin{table}[ht]
    \centering
    \begin{tabular}{rccccc}
        \toprule
        \textbf{Dataset} & \textbf{Domains} & \textbf{Data Model} & \textbf{|Eval|} & \textbf{|Test|} & \textbf{|Train|} \\
        \midrule
        ACOS \cite{acos_extract_classify} & \{Restaruant, Laptop\} & A-C-O-S & 909 & 1399 & 4464\\
        Structured Sentiment \cite{structured_sentiment_barns_2022}& Open Domain & Structured Sentiment & 2544  & 2929 & 9870\\
    
        UOC \cite{negi_uoc_2025}      & \{Laptop, Restaurant, & UOC & 100 &   N/A & N/A\\
         & Books, Clothes, Hotels\}& & &  & \\
        \bottomrule
    \end{tabular}
    \caption{Opinion Data Models}
    \label{tab:dmodel_dsets}
\end{table}
\vspace{-20pt}

\begin{itemize}
    
    \item \textbf{StockMotions} \cite{lee2023stockemotions}: It is a dataset proposed for detecting emotions in the stock market that consists of 10k English comments collected from StockTwits, a financial social media platform. This dataset is annotated for 2 sentiments (bearish, bullish) and twelve emotions (ambiguous, amusement, anger, anxiety, belief, confusion, depression, disgust, excitement, optimism, panic and surprise). This dataset is sourced from a finance specific social media platform known as stocktwits.
    \item \textbf{Political News Fact Checking} \cite{misra2022not}: This dataset contains 21,152 high-quality fact-checked statements along with their sources (medium, broadcasting channels, social media platforms etc). All statements are classified into one of 6 categories: \textit{true, mostly true, half true, mostly false, false and ``pants on fire"}.
   
\end{itemize}
\begin{table}[ht]
    \centering
    % \large    
    \begin{tabular}{rccccc}
        \toprule
        \textbf{Dataset} & \textbf{From} & \textbf{To} & \textbf{Total} & \textbf{Unique} & \textbf{Daily Median} \\
        \midrule
        Stockmotions \cite{lee2023stockemotions} & 2020-01-21 & 2020-12-31 & 10\,000 & 1\,341  & 24\\
        Politifact     \cite{misra2022not}      & 2000-10-01 & 2022-07-09 & 21\,152  & 4\,751 & 4\\
        \bottomrule
    \end{tabular}
    \caption{Temporal coverage and size of the knowledge-base dataset}
    \label{tab:tdatasets}
\end{table}
\vspace{-10pt}

\section{Methods}

\subsection{Overview}
We utilise the data models (\ref{sec:data_models}) to enrich the existing classification datasets (\ref{tab:tdatasets}) by annotating it for the subjective and contextual opinion information expressed in the text. This annotation extension adds a more fine-grained semantic decomposition of the opinions expressed in the text, aiding a more comprehensive analysis. 
\begin{figure}[h]
    \centering
    % Include the PDF as a graphic
    \includegraphics[width=\linewidth]{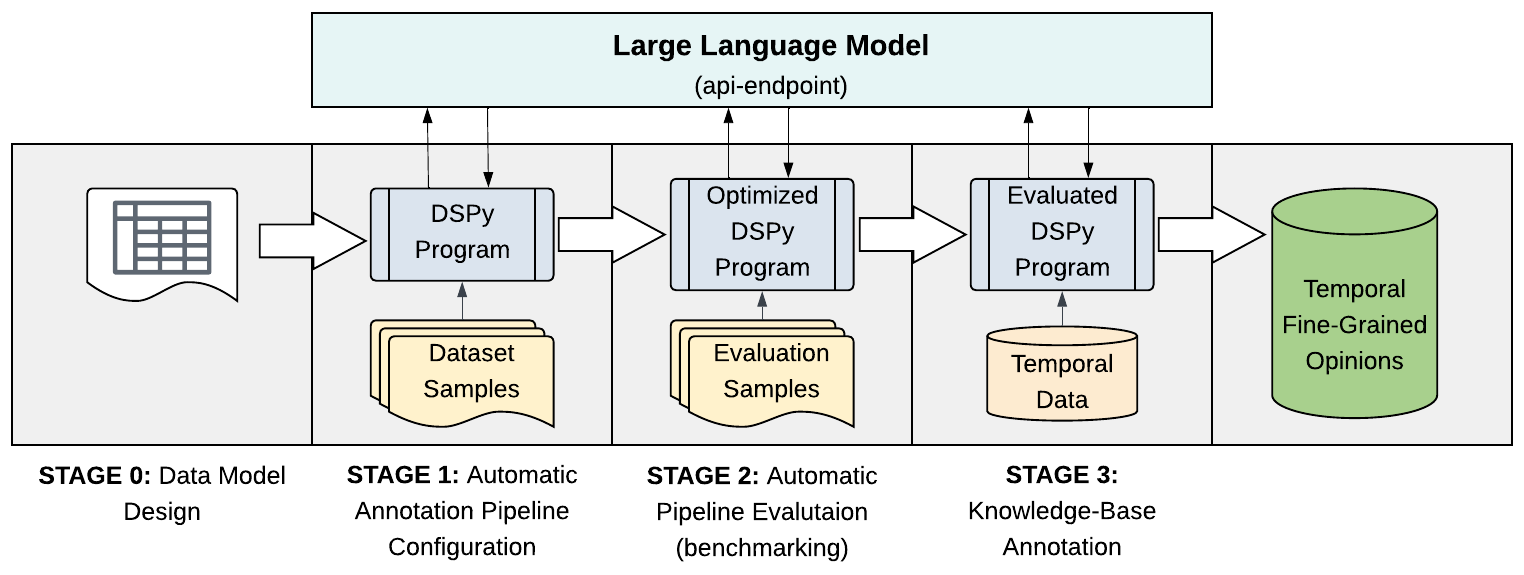}
    \caption{Flow diagram representing the DSPy-LLM architecture and knowledge generation pipeline.}
    \label{fig:kbg-diagram}
\end{figure}
The proposed annotation pipeline uses LLMs for automatic annotation. However, LLMs are known to be sensitive to the prompts \cite{zhuo-etal-2024-prosa} and exhibit a higher level of variability in performing complex tasks \cite{llm_reproduce_2025}. To minimise the spurious interaction between the prompt and model selection as a confounding variable, we adopt \textbf{DSPy} \cite{dspy_2024} method of programming LLMs rather than prompting them naively. Figure \ref{fig:kbg-diagram} illustrates the three-step annotation pipeline for creating a knowledge-base of Temporal Fine-Grained opinions. 

\begin{enumerate}
    \item \textbf{Automatic Annotation Pipeline Configuration and Training}: In the first stage, we configure the DSPy Program, which includes the selection of the optimum prompts and examples for In-Context Learning (ICL) for the interaction with the LLM(s). The DSPy Program automatically carries out this stage with the help of a small dataset sample, data model and evaluation metric, which enables the bootstrapping of the ICL examples to the prompt. The output generated by the LLM is marshalled into the predefined data model by the DSPy program. 
    \item \textbf{Pipeline Evaluation}: This is the evaluation stage of our automatic annotation pipeline. We measure the performance of the annotation pipeline (DSPy + LLM) on the test sample of the annotated opinion mining dataset. It informs us of the efficacy of our automatic annotation process.
    \item \textbf{Temporal Opinion Annotation}: In the final stage of our pipeline, we ingest timestamped opinionated data into our automatic annotation pipeline, which in turn generates annotations to create a temporal knowledge base for opinions extracted from social media posts.
\end{enumerate}

% \subsection{Prompt Engineering}
% \paragraph{A big question to answer if justification for our prompt and prompting methods?}

\subsection{Dataset Sampling Protocol for Training and Evaluation of the Annotation Pipeline}
\label{sec:sampling}
The annotation pipeline relies on the ability of pre-trained language models to annotate opinions. However, we still utilise the available annotated datasets to: i) Provide DSPy programs with examples for ICL, essentially for training our pipeline. It is worth noting that this does not involve training the weights of the LLM. (ii) Evaluate DSPy programs for the best performing program configurations for LLM interaction. 

Our annotation pipeline is LLM-driven, and thus testing and configuring DSPy programs is computationally expensive. To make the program configuration computationally optimal, we comprehensively sample the annotated dataset for ease of DSPy programming and pipeline evaluation. 

\paragraph{Outlier Estimation} Prior to the sampling process, we analyse the distribution of the number of opinion labels per text example in the annotated datasets to identify outliers. We exclude the outliers because they represent the cases where the number of annotated opinions is exceptionally high or exceptionally low. We use the instances from the dataset as examples for ICL, thus serving as a template for the expected annotation output in our pipeline. Therefore, it becomes important to exclude the outlier instances as they could skew the decision of our LLM-based annotation pipeline when deciding how many opinions to annotate for each text instance.

We perform outlier exclusion on both the test and training datasets for a practical training and evaluation process.  

\begin{figure}[ht]
  % \centering
  \centering

 \includegraphics[width=\linewidth]{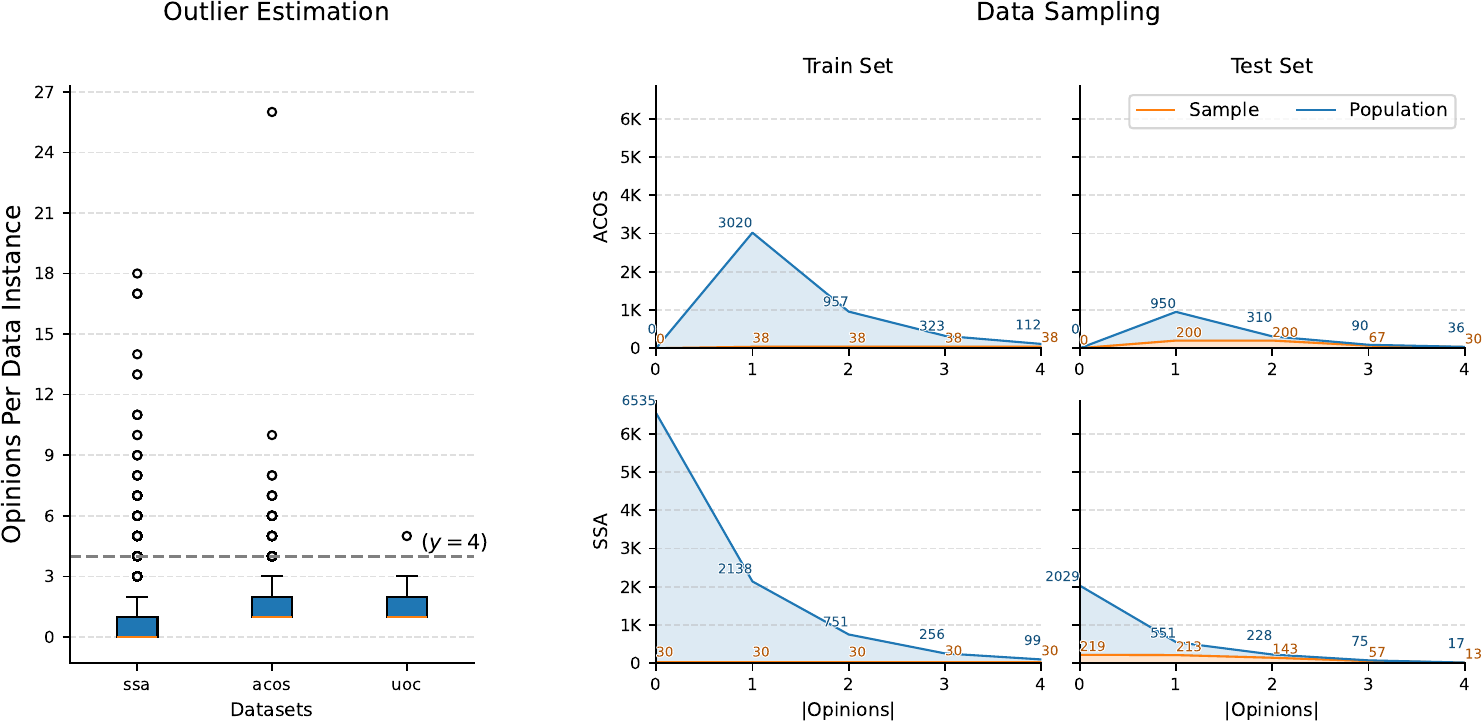}

  \caption{Opinion per-data instance (Left), Original And Sampled Datasets (Right)}
  \label{fig:sampling_comparison}
\end{figure}

Figure. \ref{fig:sampling_comparison} (Right) compares the distribution of the number of opinions annotated across three datasets associated with the data models. It also shows our upper bound for the exclusion criteria and decides which opinion densities are considered outliers. We calculate the upper bounds ($U_{dataset}$) by using the estimated inter-quartile ranges after combining all the dataset splits (train, validation and test). To calculate individual upper bounds, we utilise the interquartile ranges ($IQR$) to determine the upper bound. $$U_{dataset} = Q_3 + 1.5 \times IQR$$ The individual upper bound values are $(2.5, 4, 4)$ for $(U_{ssa}, U_{acos}, U_{uoc})$ respectively. We select the combined upper bound of four using $ \max(2.5, 4, 4) = 4$.

Figure. \ref{fig:sampling_comparison} (Left) shows the distribution of the train and test sets of the datasets. We select small, equal numbers of inlier examples from the training set using stratified random sampling. We do this for the datasets having a substantial amount of data (SSA, ACOs), as we want those examples for DSPy programs to select ICL examples. For testing, in addition to stratified random sampling, we also preserve an element of label density; however, we do reduce the overrepresented cases.

\subsection{Evaluation Metric}
\label{sec:metric}
The evaluation metrics take the agreement with the ground truth across the extracted opinion tuples and also the agreement with the ground truth of individual elements of extracted opinions. The Tuple-level exact match metric severely penalizes the mismatch in the measured values; even a slight mismatch of one component completely devalues the entire extracted opinion. In doing so, it does not account for the partially correct extracted opinions, exacerbating the non-linearity or discontinuity of the evaluation metrics discussed in elaborate detail by \citet{nemerge_2023}. Therefore, our metric of choice is the component-level exact match metric discussed in the remainder of this section.

In the dataset with text instances $\{T_i\}_{i=1}^N$ for each text instance $T_i$ there exists the ground truth opinion annotation $Og_i$ is a set of opinions $Og_i = \{og_{i,j}|j=1,2,...,|Og_i|\}$ and the corresponding set of predicted opinions $Oe_i=\{oe_{i,k}|k=1,2,...|Oe_i|\}$. For any pair of tuples $(oe_{i,k}, og_{i,j})$ we describe the degree of agreement as:

$$f(oe_{i,k}, og_{i,j}) = \frac{|oe_{i,k} \cap og_{i,k}|}{|og_{i,k}|}$$
We perform a one-to-one matching (without replacement) between the tuples in $Oe_i$ and $G_i$. Now $\mathcal{A}_i \subseteq Og_{i} \times Oe_{i}$, is the set of aligned tuple pairs obtained. For each gold tuple $og_i\in G_i$ at most one predicted/extracted tuple is selected (without replacement, one predicted tuple cannot be matched with other ground truth tuples.). The selection can also be shown as:
$$\mathcal{A}_i = \arg \max_{\mathcal{M}\subseteq Og_i \times Oe_i \text{matching}} \sum_{og,oe \in \mathcal{M}} f(oe, og)$$
Any extracted tuple not included in $\mathcal{A}_i$ does not contribute towards true positive. However, it does bring precision down as it is considered when counting the total extracted opinion tuples. Now for each text input $T_i$ we calculate true positive $$TP= \sum_{i=1}^{N} \sum\limits_{(og,oe)\in \mathcal{A}_i}f(oe, og)$$ 
Precision $P$ and recall $R$ and F1 score are then given by:
$$P=\frac{TP}{\sum_{i=1}^{N} |Oe_i|} \text{ , } R= \frac{TP}{\sum_{i=1}^{N}|Og_i|} \text{ , } F1 = \frac{2 * P * R}{P + R}$$

\subsection{Annotation Pipeline Training and Configuration}
We use the DSPy framework for all our interactions with LLMs. It shifts focus from performing string manipulation of the prompt strings to programming with structured and declarative natural-language modules. However, in order to train and configure the pipeline, we require the following:
\begin{enumerate}[label=\roman*., nosep]
    \item \textbf{DSPy Signature}: A Signature defines the input (text) and output (data model) specification for a single sub-task within a program. 
    \item \textbf{Program}: A Program that is a composition of one or more signatures into a logical pipeline. The program in our pipeline is configured for the annotation following the data models previously defined (Section \ref{sec:data_models}).
    \item \textbf{Evaluation function}: An evaluation function used to supervise the optimisation of a program. The evaluation function in this case implements the evaluation metrics described in Section \ref{sec:metric}.
    \item \textbf{Optimiser}: We use MIPRO optimiser \cite{mipro2024} for training the annotation pipeline. It synthesises the prompt and selects the most useful ICL examples, optimising the annotation pipeline's interaction with LLMs. 
    \item Annotated Sample: Training sample for optimisation of the DSPy program. The sampling methods described in Section \ref{sec:sampling} are applied to get training and evaluation samples from the datasets described in the Table. \ref{tab:dmodel_dsets}.
    \item \textbf{Compiler}: Binds a program with training data and a metric to learn internal prompt structures.
\end{enumerate}

\subsection{Evaluation of Annotation Process}
Once we have configured the annotation pipeline, we perform a two-step evaluation of our pipeline. For both assessments, we report the precision, recall, and F1 scores. 
\begin{enumerate}
    \item \textbf{Pipeline Efficacy Evaluation:} The evaluations are performed on the test sample derived from the pipeline training dataset. This evaluation is against human-annotated labels, with our objective being the measurement of adherence of LLM annotations to human ones. We also use these results to select the best-performing settings for applying them to the annotation of the temporal opinion datasets. These results are reported in the Table. \ref{tab:annotation_pipe_results}.
    \item \textbf{Annotation Consistency Evaluation:} Once the temporal knowledge base has been annotated, we evaluate the LLM annotations against each other. The objective of this evaluation is to measure agreement between two artificial annotators (based on LLMs). Since these are structured and span-based expression level assessments, we follow the suite of accepted methodologies in opinion mining and semantic role labelling \cite{toprak_2010, wilson_etal_annot_2005, acos_extract_classify} by using the F1 score between the annotations of two annotators as a proxy for IRR. The annotator agreement results are reported in the Table. \ref{tab:annotation_agree}. 
\end{enumerate}

\section{Experiment Setup}
We conduct the experiments with Ministral-8B \cite{llm_mistral_7b} \footnote{mistralai/Ministral-8B-Instruct-2410} and Llama-3.1-8B \cite{llm_llama} \footnote{meta-llama/Llama-3.1-8B-Instruct}. The models were hosted locally using SGLang \footnote{https://github.com/sgl-project/sglang} serving framework as required by DSPy for  OpenAI compliant API access to LLMs for prompt optimisation and inference. We keep the LLM hyperparameters constant across all models. The temperature is set to 0.0 to ensure the most deterministic generation. The context window of the model determines the input sequence length; it is 128K for both models. The output length of the generated sequence is set to 4096. All the experiments were conducted on a machine with one NVIDIA GeForce RTX 4090 (24 GB GPU memory).

We train the annotation pipeline on a different number of training samples based on the availability of the training data. For ACOS and SSA annotation, where the human-annotated data is ample, we provide 150 and 152 examples to the DSPy compiler, respectively, to configure the prompt and select ICL examples. In the case of UOCE,  the annotated data sample is scarce, so we only use 30 examples at this stage. The different DSPy configurations we test are the inclusion of 0, 5, 10 and 15 ICL examples with and without Chain-Of-Thought (COT) reasoning \cite{cot_2022} as an intermediate stage. 

\section{Analysis}
\subsection{Quantitative Analysis}

\begin{table}[ht]
\center
\resizebox{\textwidth}{!}{%
\begin{tabular}{l l c ccc ccc ccc}
% \toprule
\rowcolor{gray!15}\textbf{Model} & \textbf{Setting} & \textbf{COT} &
\multicolumn{3}{c}{\textbf{SSA}} &
\multicolumn{3}{c}{\textbf{ACOS}} &
\multicolumn{3}{c}{\textbf{UOC}} \\
 \rowcolor{gray!15}&  & (Y/N) & \textbf{Prec.} & \textbf{Rec.} & \textbf{F1} & \textbf{Prec.} & \textbf{Rec.} & \textbf{F1} & \textbf{Prec.} & \textbf{Rec.} & \textbf{F1} \\
\midrule
\multirow{8}{*}{Llama-3.1-8B}
  & \multirow{2}{*}{Zero-Shot} & N & 19.71 & 30.84 & 24.05 & 20.61 & 26.54 & 23.20 & 28.57 & 37.20 & 32.32 \\
  &  & Y & 30.44 & 27.20 & 28.73 & 30.26 & 24.21 & 26.90 & 40.72 & 39.30 & 39.99 \\
  % \cmidrule(rr){2-12}
  & \multirow{2}{*}{MIPRO (5-shot)} & N & 28.75 & 43.66 & 34.67 & 49.16 & 50.12 & 49.63 & 49.81 & 60.23 & 54.53 \\
  &  & Y & 45.03 & 42.92 & 43.95 & 47.67 & 46.58 & 47.12 & 58.35 & 57.67 & \textbf{58.01} \\
  % \cmidrule(rr){2-12}
  & \multirow{2}{*}{MIPRO (10-shot)} & N & 31.27 & 47.63 & 37.75 & 51.19 & 60.02 & 55.25 & 46.30 & 64.07 & 53.76 \\
  &  & Y & 44.91 & 46.96 & \underline{\textbf{45.91}} & 54.76 & 52.92 & 53.83 & 50.11 & 53.60 & 51.80 \\
  % \cmidrule(rr){2-12}
  & \multirow{2}{*}{MIPRO (15-shot)} & N & 39.74 & 48.33 & 43.61 & 57.25 & 62.84 & \underline{\textbf{59.92}} & 43.78 & 60.58 & 50.83 \\
  & & Y & 34.61 & 46.07 & 39.52 & 54.93 & 58.57 & 56.69 & 56.15 & 59.42 & 57.74 \\
\midrule
\multirow{8}{*}{Mininstral-8B}
  & \multirow{2}{*}{Zero-shot} & N & 26.43 & 21.87 & 23.93 & 31.14 & 23.53 & 26.81 & 39.51 & 38.13 & 38.81 \\
  
  &  & Y & 33.21 & 23.87 & 27.78 & 30.31 & 19.78 & 23.94 & 41.36 & 35.12 & 37.99 \\
  % \cmidrule(rr){2-12}
  & \multirow{2}{*}{MIPRO (5-shot)} & N & 32.05 & 48.88 & 38.72 & 47.54 & 57.65 & 52.11 & 52.57 & 61.74 & 56.79\\
  &  & Y & 43.56 & 44.58 & 44.06 & 46.03 & 38.48 & 41.92 & 61.15 & 55.46 & \underline{\textbf{58.17}}\\
  % \cmidrule(rr){2-12}
  & \multirow{2}{*}{MIPRO (10-shot)} & N & 42.08 & 49.97 & \textbf{45.69} & 56.61 & 56.61 & 56.61 & 49.36 & 63.13 & 55.4 \\
  &  & Y & 39.24 & 43.79 & 41.39  & 48.89 & 47.67 & 48.27 & 50.28 & 62.55 & 55.75 \\
  % \cmidrule(rr){2-12}
  & \multirow{2}{*}{MIPRO (15-shot)} & N & 35.59 & 49.17 & 41.29  & 48.21 & 64.8 & 55.28 & 50.93 & 63.37 & 56.48\\
  &  & Y &  42.25 & 45.67 & 43.89 & 60.32 & 56.46 & \textbf{58.33} & 55.46 & 55.46 & 55.46 \\
\bottomrule
\end{tabular}%
}
\caption{Performance comparison across Structure Sentiment Analysis Test Sampling (SSA), ACOS, and UOC datasets using \textit{Llama-3.1-8B} and \textit{Ministral-8B} under various prompting strategies.}
\label{tab:annotation_pipe_results}
\end{table}

\paragraph{Performance on human-annotated test sample.} These results are reported in Table \ref{tab:annotation_pipe_results}. 
We observe that the best-performing setting in terms of the number of included ICL examples remains consistent across both Ministral-8B and Llama-3.1-8B when viewed within the same data model annotation evaluation. For SSA, we get the best f1-score of \underline{\textbf{45.91}} with Llama-3.1-8B, with 10 ICL examples and COT applied, closely followed by Mininstral-8B at \textbf{45.69}, also 10 ICL examples but with no COT applied. When evaluating the F1-score for ACOS annotation, the best performance in our experiments is by Llama-3.1-8B at \underline{\textbf{59.92}} with 15 ICL examples and no COT applied, the close second is again 15 ICL examples by Ministral-8B and with COT applied, it gets an F1-score of \textbf{58.33}.
In the case of UOCE annotations evaluation, Ministral-8B gets an f1-score of \underline{\textbf{58.17}} with 5 ICL examples with COT applied, closely followed by Llama-3.1-8B at \textbf{58.01} also with 5 ICL examples included and COT applied.

The inclusion or exclusion of COT reasoning does not seem to have a definitive effect on the scope of our models and experiments. Only in the Zero-Shot setting, on average, the results seem to improve with COT reasoning applied.

\paragraph{Measurement of annotation agreement across LLMs}

\begin{table}[b]
\centering
\begin{tabular}{l|c| c c c|c c c}
\hline
\multicolumn{2}{c|}{Datasets $\rightarrow$} & \multicolumn{3}{c|}{\textbf{Politifacts}} & \multicolumn{3}{c}{\textbf{Stockmotions}} \\
\hline
\multicolumn{2}{c|}{Data Models $\rightarrow$} & SSA & ACOS & UOCE & SSA & ACOS & UOCE \\
\hline
\multirow{11}{*}{\rotatebox[origin=c]{90}{Opinion Concepts}} & Target & 37.27 & -- & -- & 23.71 & -- & -- \\
% \cline{2-8}
 
  & Holder Entity & -- & -- & 27.58 & -- & -- & 78.23 \\

 & Reason & -- & -- & 90.97 & -- & -- & 93.64 \\
 
 & Qualifier & -- & -- & 96.49 & -- & -- & 94.42 \\
 & Senitment Intensity & 51.62 & -- & 58.01 & 33.12 & -- & 71.23 \\
 & Holder Span & 70.88 & -- & 56.42 & 67.83 & -- & 25.86 \\
 & Aspect Term & -- & 30.59 & 26.29 & -- & 37.72 & 14.43 \\
 & Entity & -- & 14.85 & 21.07 & -- & 20.74 & 18.25 \\
 & Category & -- & 24.5 & 45.26 & -- & 40.53 & 59.15 \\
 & Setiment Polarity & 67.67 & 73.39 & 76.52 & 47.61 & 71.24 & 78.45 \\
 & Sentiment Expression* & 39.4 & 20 & 42.98 & 21.61 & 26.5 & 30.92 \\
\hline
\end{tabular}
\caption{Inter-LLM annotation agreement measured using span-level F1 }
\label{tab:annotation_agree}
\end{table}

Since we are annotating the temporal subjective dataset with opinions, we do not have human-annotated labels. In order to overcome this limitation, we measure annotation agreement between the opinions annotated by Ministal-8B and Llama-3.1-8B models. We select the best-performing annotation pipeline setting for both LLMs across all data models. 

We follow the suite of accepted methodologies in opinion mining and semantic role labelling \cite{toprak_2010} by using the F1 score between the annotations of two annotators as a proxy for IRR. Even our agreement scores follow a similar trend as they observed in the results reported in Table \ref{tab:annotation_agree}. 
We see that opinion components that represent classification labels, such as Sentiment Polarity and Sentiment Intensity, have a higher degree of agreement across both LLMs and all data models. The component where more subjective expressions are possible, like entity, also has low overall scores.

The LLM annotations for each data model show high consistency of label agreement for both the most and least agreeable values across both datasets. As one would expect, the SSA data models specialising in capturing structured sentiments agreeable on holder spans, sentiment polarity. Across all data models capturing aspect terms (least agreeable opinion concept), the ACOS data model leads to aspect term extraction yielding the most agreement in both LLMs. UOCE shows the highest agreement for Sentiment Intensity, Sentiment Expression, Sentiment Polarity and Aspect Category. However, the aspect terms annotated by it show high disagreement compared to the aspect terms annotated when the ACOS data model is used. We believe that these results are promising and would only improve with LLMs having a higher number of model parameters.

\begin{table}[b]
    \centering
    \scriptsize
    \begin{tabular}{p{2.5cm}|>{\raggedright\arraybackslash}p{4cm}|p{4cm}|p{4cm}}
    \toprule
    \textbf{DATA POINTS} & \textbf{SSA} & \textbf{ACOS} & \textbf{UOCE} \\
    \hline
     ID: \newline POLITIFACT\_00000 \newline DATE: 2008-06-11 \newline POST: John McCain opposed bankruptcy protections for families "who were only in bankruptcy because of medical expenses they couldn\'t pay."&  \begin{lstlisting}
[{'sentiment': {
'polarity':'negative',
'expression':'opposed',
'intensity':'average'
},
'target': 'bankruptcy 
protections',
'holder': 'John McCain'
 }]
\end{lstlisting}&  \begin{lstlisting}
[{'sentiment': 'Negative',
  'aspect_term': 'bankruptcy 
  protections',
  'aspect_category': {
    'entity': 'bankruptcy 
    law',
    'entity_attribute': 
    'general'
  },
  'opinion_span': None
 },
 {'sentiment': 'Negative',
  'aspect_term': 'families',
  'aspect_category': {
    'entity': 'people',
    'entity_attribute': 
    'general'
  },
  'opinion_span': None
 },
 {'sentiment': 'Negative',
  'aspect_term': 'medical 
  expenses',
  'aspect_category': {
    'entity': 'healthcare',
    'entity_attribute': 
    'cost'
  },
  'opinion_span': None
 }]
\end{lstlisting}&  \begin{lstlisting}
[{'entity': 'Policy',
  'holder': {
    'holder_span': 'John 
    McCain',
    'holder_entity': 'John 
    McCain'
  },
  'aspect': {
    'term': 'bankruptcy 
    protections',
    'category': 'General'
  },
  'sentiment': {
    'polarity': 'positive',
    'expression': None,
    'intensity': 'weak'
  },
  'reason': 
  'sympathetic to 
  families in 
  bankruptcy due 
  to medical expenses',
  'qualifier': 
  "families who 
  were only in 
  bankruptcy 
  because of medical 
  expenses they 
  couldn't pay"
 }]
\end{lstlisting}
 \\
    \hline
    \end{tabular}
    \caption{Example of Annotation instance}
    \label{tab:egannot}
\end{table}

\subsection{Example of Annotated Instances}
The example of an annotated instance for the dataset is displayed in the Table \ref{tab:egannot}. It shows how our annotation pipeline annotates different facets of opinion in the three defined data models. It can be understood from this instance how the agreement might vary when the model has a high degree of freedom when picking up categories, entities and when selecting the spans from the text input in question. 

E.g. ACOS data model recognises multiple entities for this instance, which include \textit{``bankruptcy law'', ``people'', ``healthcare''}. In contrast, the UOCE formulation, being more expressive, combines them at a higher level and recognises the entity \textit{``Policy''}. However, only SSA can capture the sentiment expression \textit{``oppose''} that expresses the correctly recognised \textit{``Negative''} sentiment.

\section{Conclusion}
We present a scalable framework for building a temporal opinion knowledge base by leveraging LLMs as automated annotators. Our approach bridges the gap between fine-grained opinion mining and temporally structured social media analysis by incorporating established opinion schemas within a declarative annotation pipeline. The framework eliminates the need for manual prompt engineering when annotating fine-grained opinions with LLMs. We define three data models that guide the representation of temporally grounded sentiment and opinion components. To assess annotation quality, we conduct a rigorous evaluation using human-annotated samples and compute annotator agreement between two LLMs across each opinion dimension, following established human annotation practices. 

\section{Limitations and Future Work}
The primary focus of this work is on the methodology for creating a knowledge base of temporal subjective data by annotating it for opinions. The work focuses on utilising LLMs in a declarative manner and takes into account accepted opinion formulations, which we utilise as the data models. Yet several limitations exist that must be acknowledged. Firstly, we only used open-weight LLMs that are considered "small" in size, having 8 billion parameters each. Even though the utilised LLMs have shown acceptable performance, scaling up the LLMs would improve performance significantly, but it was outside the scope of our current work. Secondly, we only experiment with freely available open-weight LLMs; however, the state-of-the-art proprietary LLMs are not included in this study. The existing pipeline can accommodate both these experiments without any change to the annotation setup.

Lastly, we studied the degree of consensus in the annotations by Llama-3.1-8B and Ministral-8B-Instruct for each label across three opinion data models. However, there is an important need to extend the method to ensemble the annotations in a meaningful way. We seek to answer the question of ensembling in our future work.

\section{Acknowledgments}
This work was conducted with the financial support of the Science Foundation Ireland (SFI) under Grant Number SFI/12/RC/2289\_P2 (Insight\_2). At the time of this publication, Omnia Zayed has been supported by Taighde Éireann – Research Ireland for the Postdoctoral Fellowship award GOIPD/2023/1556 (Glór)

%% The declaration on generative AI comes in effect
%% in Janary 2025. See also
%% https://ceur-ws.org/GenAI/Policy.html
% \section*{Declaration on Generative AI}
%   {\em Either:}\newline
%   The author(s) have not employed any Generative AI tools.
%   \newline
  
%  \noindent{\em Or (by using the activity taxonomy in ceur-ws.org/genai-tax.html):\newline}
%  During the preparation of this work, the author(s) used X-GPT-4 and Gramby in order to: Grammar and spelling check. Further, the author(s) used X-AI-IMG for figures 3 and 4 in order to: Generate images. After using these tool(s)/service(s), the author(s) reviewed and edited the content as needed and take(s) full responsibility for the publication’s content. 

\bibliography{samplebibliography}

\end{document}